\documentclass[a4paper,11pt]{article}

\usepackage[nohead,margin=2cm]{geometry}
\usepackage{framed,multirow}

\usepackage{amssymb}
\usepackage{latexsym}
\usepackage{multirow}
\usepackage{framed,multirow}
\usepackage{amsmath}
\usepackage{epsfig}
\usepackage{setspace}
\usepackage{url}
\usepackage{xcolor}
\usepackage{lscape} 
\usepackage{authblk}
\graphicspath{{figures/}}

\begin{document}

\title{Survey on Reliable Deep Learning-Based Person Re-Identification Models: Are We There Yet?}

\author[1]{Bahram Lavi\thanks{Corresponding author: bahram.lavi@ic.unicamp.br}}
\author[2]{Ihsan Ullah}
\author[3]{Mehdi Fatan}
\author[1]{Anderson Rocha}
\affil[1]{\footnotesize Institute of Computing, University of Campinas (UNICAMP), Campinas, S\~ao Paulo, Brazil.}
\affil[2]{\footnotesize Data Mining \& Machine Learning Group, Discipline of IT, National University of Ireland Galway, Ireland.}
\affil[3]{\footnotesize Department of Computer Engineering and Mathematics, University Rovira i Virgili, Tarragona, Spain}

\date{}

\maketitle

\begin{abstract}
\emph{Intelligent video-surveillance} (IVS) is currently an active research field in computer vision and machine learning and provides useful tools for surveillance operators and forensic video investigators.
Person re-identification (PReID) is one of the most critical problems in IVS, and it consists of recognizing whether or not an individual has already been observed over a camera in a network. Solutions to PReID have myriad applications including retrieval of video-sequences showing an individual of interest or even pedestrian tracking over multiple camera views. Different techniques have been proposed to increase the performance of PReID in the literature, and more recently researchers utilized deep neural networks (DNNs) given their compelling performance on similar vision problems and fast execution at test time. Given the importance and wide range of applications of re-identification solutions, our objective herein is to discuss the work carried out in the area and come up with a survey of state-of-the-art DNN models being used for this task. We present descriptions of each model along with their evaluation on a set of benchmark datasets. Finally, we show a detailed comparison among these models, which are followed by some discussions on their limitations that can work as guidelines for future research. 
\end{abstract}

\section{Introduction}
\label{sec:introduction}
The importance of security and safety of people in society at large is continuously growing. Governmental and private organizations are seriously concerned with the security of public areas such as airports and shopping malls. It requires significant effort and financial expense to provide security to the public. To optimize such efforts, video surveillance systems are playing a pivotal role. Nowadays, a panoply of video cameras is growing as a useful tool for addressing various kinds of security issues such as forensic investigations, crime prevention, and safeguarding restricted areas. 
\par
Daily continuous recording of videos from network cameras results in daunting amounts of videos for analysis in a manual video surveillance system. Surveillance operators need to analyze them at the same time for specific incidents or anomalies, which is a challenging and tiresome task. Intelligent video surveillance systems (IVSS) aim to automate the issue of monitoring and analyzing videos from camera networks to help surveillance operators in handling and understanding the acquired videos. This makes the IVSS area one of the most active and challenging research areas in computer engineering and computer science for which computer vision (CV) and machine-learning (ML) techniques plays a key role. This field of research enables various tools such as ~\textit{online} applications for people/object detection and tracking, recognizing suspicious action/behavior from the camera network; and 
~\textit{off-line} applications to support operators and forensic investigators to retrieve images of the individual of interest from video frames acquired on different camera views. 

Person identification is one of the problems of interest in IVSS. It consists of recognizing an individual over a network of video surveillance cameras with possibly non-overlapping fields of view~\cite{bedagkar2014survey,saghafi2014review, iodice2016strict}. 
In general, the application of PReID is to support surveillance operators and forensic investigators in retrieving videos showing an individual of interest, given an image as a query (a.k.a. \textit{probe}). Therefore, video frames or tracks of all the individuals (a.k.a.~\textit{template gallery}) recorded by the camera network are sorted in descending order of similarity to the probe. It allows the user to find occurrences (if~\textit{any}) of the individual of interest in the top positions. 
\par
Person re-identification is a challenging task due to low-image resolution, unconstrained pose, illumination changes, and occlusions, which adhere to the use of robust biometric features like face, among others.
Whereas, some cues like gait and anthropometric measures have been used in some existing PReID systems.
Most of the existing techniques rely on defining a specific descriptor of clothing (typically including color and texture), and a specific similarity measure between a pair of descriptors (evaluated as a~\textit{matching score}) which can be either manually defined or learned directly from data~\cite{bedagkar2014survey,gray2008viewpoint,farenzena2010person,hirzer2012person,ma2014covariance}.

%
\par
\begin{figure}[!t]
\centering
\includegraphics[width=0.7\textwidth]{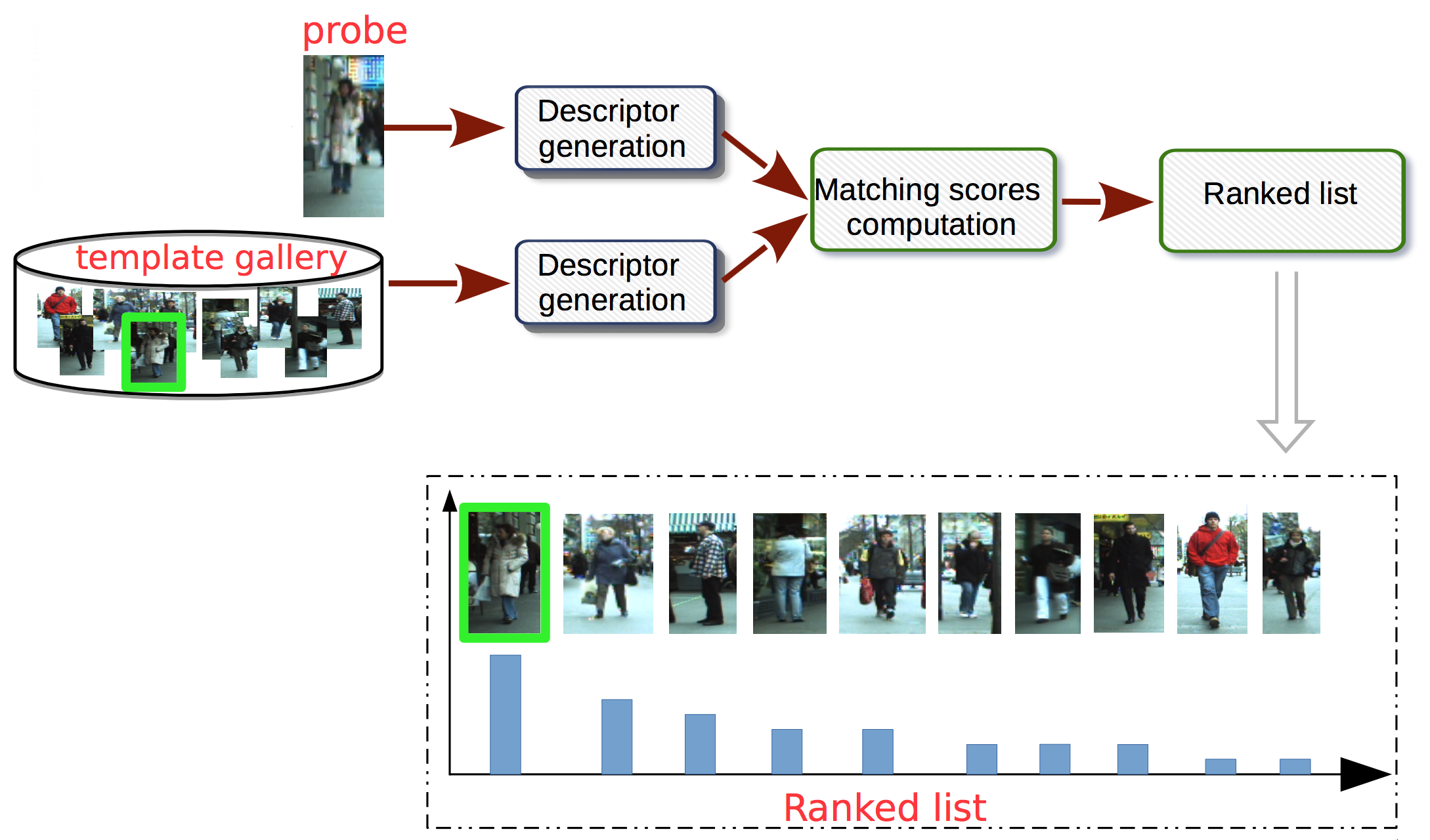}
\caption{Standard person re-identification system. Given a probe image and set of template images, the goal is to generate a robust image signature from each and compute the similarity between them, and finally presented by a sorted, ranked list.}
  \label{fig:standard_re_id}
\end{figure}

\textbf{Standard PReID Methodology}: For a given image of an individual (a.k.a.~\textit{probe}), a PReID system aims to seek the corresponding images of that person within the gallery of templates.
Take into consideration that creating the template gallery depends on a re-identification setup, which we can categorize as:
(i) \textit{single-shot} which has only one template frame per individual, and (ii) \textit{multiple-shots} that contains more than one template frame per individual.
In this case, a continuous PReID system is employed in real-time whereby the individual of interest is continuously matched against the template image with the gallery set, using the currently seen frame as a probe. 
Figure.~\ref{fig:standard_re_id} demonstrates a basic PReID framework. After an image description is generated for a probe, and the template images of the gallery set, matching scores between each of them are computed; and finally, the ranked list is generated by sorting the matching scores in decreasing order.

\par
The strategy of many existing descriptors use hand-crafted features.  Deep-learning (DL) models -- e.g.,convolutional neural networks (CNNs)~\cite{lecun2010convolutional, krizhevsky2012imagenet,russakovsky2015imagenet}-- have been particularly used to solve the problem of PReID by learning from data.
A CNN-based model generates a set of feature maps, whereby each pixel of a given image corresponds to specific feature representation. The desired output is expected at the top of the employed model. There are different approaches to train a deep neural network (DNN) model. A DNN model can be trained in a Supervised, Semi- and Un-Supervised manner depending on the problem scenarios and availability of labelled data. In the task of PReID only a small set of training data is available. Thus, developing a learning model in semi- and un-supervised manner is usually challenging task and the model might result in failure or poor performance in PReID. Most of the papers discussed at the end of this paper engage with supervised learning techniques, and only a few of them considered semi- or un-supervised approach. Further, we consider the models used for PReID in three categories as \{single, pairwise, and triplet\} feature-learning strategies. Details are presented and discussed in section~\ref{sec:deep_re_techniques}.
This paper presents the state-of-the-art methods of PReID techniques based on DNNs and provides a significant detailed information about them. The literature review involves papers which were published between the year 2014 to date. We provide a taxonomy of deep feature learning methods for PReID including comparisons, limitations, and future research directions and potential as well as opportunities for research in the horizon. Unlike~\cite{wu2019deep,leng2019survey}, we provide a comprehensive and detailed review of the existing techniques, particularly, the more modern ones that rely upon DNN feature learning strategies. We stress that, in this paper, we only consider recent DNN techniques which directly involved on the procedure of PReID task.
For each technique, we analyze its experimental result and further make comparisons of the achieved performances considering different perspectives such as comparing DNNs performances when adopting different strategies to solve a problem such as the learning strategy (e.g., single, pairwise, and triplet learnings). 
\par
The structure of this paper is organized as follows: Section~\ref{sec:datasets_reid} briefly explain the benchmark datasets employed for PreID.
Section~\ref{sec:deep_re_techniques} describes DNN methods by highlighting the impact of their important content such as objective function, loss functions, data augmentation, among others. Whereas, section \ref{sec:experimental_comparison} discusses performance measures, results and their comparisons, and their limitations and future directions. Finally, section~\ref{sec:Conclusions} concludes the paper and gives final remarks about the PReID and the paper.

\section{Person Re-identification Benchmark Data sets}
\label{sec:datasets_reid}
Data is one of the important factors for current DNN models.

Some factors must be taken into account to reach a
reliable recognition rate when evaluating person re-identification solutions. Each dataset is collected to specially target one or more of these factors. The factors that create issues for PReID task includes occlusion (apparent in i-LIDS dataset) and illumination variation (common in most of them). On the other hand, background and foreground segmentation to distinguish the person's body is a challenging task. Some of the datasets provide the segmented region of a person's body (e.g., on VIPeR, ETHZ, and CAVIAR datasets). While other datasets have been prepared to evaluate the re-identification task. The most widely datasets are VIPeR, CUHK01, and CUHK03. 
VIPeR, CAVIAR, and PRID datasets are used when only two fixed camera views are given to evaluate the performance of person re-identification methods. Table.~\ref{tab:dataset_summery} gives a summary of each dataset. Below we briefly discuss each of them. 

~\textbf{VIPeR}~\cite{gray2008viewpoint}: VIPeR is a challenging dataset due to its small number of images for each individual. It is made up of two images of 632 individuals from two camera views. It consists of pose and illumination variations. The images are cropped and scaled to $128 \times 48$ pixels. This is one of the most widely used datasets for PReID and a good starting point for new researchers in PReID. Enhancing Rank-1 performance on this dataset is still an open challenge.

\textbf{i-LIDS}~\cite{branch2006imagery}: It contains 476 images of 119 pedestrians taken at an airport hall from non-overlapping cameras with pose and lighting variations and strong occlusions. A minimum of two images and an average of four images exist for each pedestrian.  

\textbf{ETHZ}~\cite{ess2007depth}: It contains three video sequences of a crowded street from two moving cameras; images exhibit considerable illumination changes, scale variations, and occlusions. The images are of different sizes. 
This dataset provides three sequences of multiple images of an individual from each sequence.
Sequences 1, 2, and 3 have 83, 35, and 28 pedestrians, respectively. 

\textbf{CAVIAR}~\cite{cheng2011custom}: It contains 72 persons and two views in which 50 persons appear in both views while 22 persons appear only in one view. Each person has five images per view, with different appearance variations due to resolution changes, light conditions, occlusions, and different poses. 

\textbf{CUHK}: 
This dataset is divided into three distinct partitions with specific setups. \emph{CUHK01}~\cite{li2012human} includes $1,942$ images of $971$ pedestrians. It consists of two images captured in two disjoint camera views, camera (A) with several variations of viewpoints and pose, and camera (B) mainly include images of the frontal and back view of the camera.
~\emph{CUHK02}~\cite{li2013locally} contains $1,816$ individuals constructed by five pairs of camera views (P1-P5 with ten camera views). Each pair includes 971, 306, 107, 193, and 239 individuals, respectively. Each individual has two images in each camera view. This dataset is employed to evaluate the performance when the camera views in the test are different from those in training. 
Finally, \emph{CUHK03} \cite{li2014deepreid} includes $13,164$ images of $1,360$ pedestrians. This data set has been captured with six surveillance cameras. Each identity is observed by two disjoint camera views and has an average of $4.8$ images in each view; all manually cropped pedestrian images exhibit illumination changes, misalignment, occlusions, and body parts missing.

\textbf{PRID} \cite{hirzer2011person}:
This dataset is specially designed for PReID, focusing on a single-shot scenario. It contains two image sets containing 385 and 749 persons captured by camera A and camera B, respectively. The two subsets of this dataset share 200 persons in common.

\textbf{WARD}~\cite{martinel2012re}: This dataset has 4,786 images of 70 persons acquired in a real-surveillance scenario with three non-overlapping cameras having huge illumination variation, resolution, and pose changes. 

\textbf{Re-identification Across indoor-outdoor Dataset (RAiD)}~\cite{das2014consistent}: It comprise of 6,920 bounding boxes of 43 identities captured by four cameras. The cameras are categorized into four partitions in which the first two cameras are indoors while the remaining two are outdoors. Images show considerable illumination variations because of indoor and outdoor changes.

\textbf{Market-1501}~\cite{zheng2015scalable}: 
A total of six cameras are used, including 5 high-resolution cameras, and one low-resolution camera. Overlap exists among different cameras. Overall, this dataset contains 32,668 annotated bounding boxes of 1,501 identities. Among them, 12,936 images from 751 identities are used for training, and 19,732 images from 750 identities plus distractors are used for gallery set.

\textbf{MARS}~\cite{zheng2016mars}: This dataset comprises 1,261 identities with each identity captured by at least two cameras. It consists of 20,478 tracklets and 1,191,003 bounding boxes.

\textbf{DukeMTMC}~\cite{ristani2016MTMC}: This dataset contains 36,441 manually-cropped images of 1,812 persons captured by eight outdoor cameras. The data set gives access to some additional information such as full frames, frame-level ground-truth, and calibration details. 

\textbf{MSMT}~\cite{wei2018cvpr}: It consists of 126,441 images of 4,101 individuals acquired from 12 indoor and three outdoor cameras, with different illumination changes, poses, and scale variations.

\textbf{RPIfield}~\cite{zheng2018rpifield}: This dataset is constructed using 12 synchronized cameras provided by 112 explicitly time-stamped actor pedestrians through out specific paths among about 4000 distractor pedestrians.

\textbf{Indoor Train Station Dataset (ITSD)}~\cite{yuan2019multi}: This dataset has the images of people from a real-world surveillance camera captured at a railway station. It presents the image size of
$64 \times 128$ pixels and contains 5607 images, 443 identities, with different viewpoints. 

\begin{table*}[!htb]
\centering
\footnotesize 
\begin{tabular}{lcccccccl}
\hline
Dataset & Year & \shortstack{Multiple\\ images} & \shortstack{Multiple \\ camera} & \shortstack{Illumination\\ variations} & \shortstack{Pose\\ variations} & \shortstack{Partial\\ occlusions} & \shortstack{Scale\\ variations} & \shortstack{Crop image\\ size}  \\ \hline
VIPeR & 2007 & $\times$ &\checkmark& \checkmark &\checkmark &\checkmark & $\times$ &$128\times48$  \\
ETHZ &2007 & \checkmark & $\times$ & \checkmark &$\times$ &\checkmark & \checkmark & vary \\ 
PRID & 2011 & $\times$ &\checkmark &\checkmark &\checkmark &\checkmark & $\times$ & $128\times64$      \\     
CAVIAR & 2011 &\checkmark &\checkmark &\checkmark &\checkmark &\checkmark & \checkmark & vary \\
WARD &2012 & \checkmark & \checkmark & \checkmark & \checkmark & $\times$ & $\times$ &$128\times48$  \\
CUHK01 & 2012 & \checkmark &\checkmark & \checkmark &\checkmark &\checkmark& $\times$ &$160\times60$ \\ 
CUHK02 & 2013 & \checkmark &\checkmark &  \checkmark & \checkmark &\checkmark & $\times$ &$160\times60$\\ 
CUHK03 &2014 &\checkmark &\checkmark & \checkmark &\checkmark &\checkmark& $\times$ &vary\\ 
i-LIDS & 2014 &\checkmark &\checkmark &\checkmark &\checkmark &\checkmark &  \checkmark &vary\\ 
RAiD &2014 & \checkmark & \checkmark & \checkmark & \checkmark & $\times$ & $\times$ &$128\times64$    \\
Market-1501 &2015 &\checkmark &\checkmark &\checkmark &\checkmark &\checkmark & \checkmark & $128\times64$ \\ 
MARS &2016& \checkmark & \checkmark & \checkmark & \checkmark & \checkmark & $\times$ & $256\times128$ \\
DukeMTMC & 2017 & \checkmark & \checkmark &  \checkmark &  \checkmark & $\times$ & \checkmark &vary \\
MSMT & 2018& \checkmark & \checkmark & \checkmark & \checkmark & $\times$ & \checkmark &vary\\
RPIfield & 2018 & \checkmark & \checkmark & \checkmark & \checkmark & \checkmark & \checkmark &vary\\
ITSD & 2019 &  \checkmark &  \checkmark & $\times$ &  \checkmark &  \checkmark & $\times$ & $64 \times 128$\\
\hline
\end{tabular}
\caption{Summary on benchmark PReID datasets.}
\label{tab:dataset_summery}
\end{table*}

\section{Deep Neural Networks for PReID}
\label{sec:deep_re_techniques}
\begin{figure}[!ht]
\centering
\includegraphics[width=0.7\textwidth]{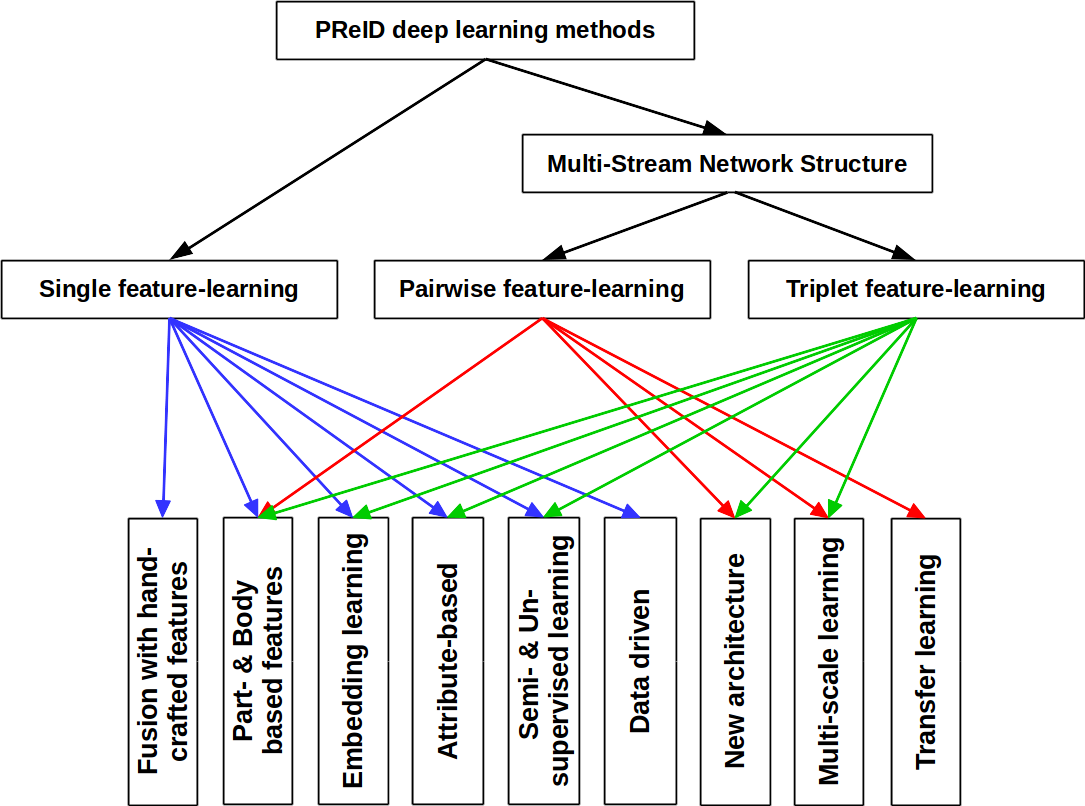}
\caption{Taxonomy of deep feature-learning methods for PReID}
  \label{fig:preid_dl_methods}
\end{figure}

 Deep learning techniques has been widely applied in several CV problems. This is due to the discriminative and generalization power of these learned models that results in promising performance and achievements. PReID is one of the challenging tasks in the area of CV for which DL models are one of the current best choice in research community. 
In the following section, we provide an overview of recent DL works for the task of PReID. Several interesting DL models have been proposed to improve PReID performance. These state-of-the-art DL approaches can be categorized by taking into account the learning methodology of their models that have been utilized in the PReID systems. Some works consider the PReID as a standard classification problem. On the other hand, some works have considered the issue of lack of training data samples in the PReID task and proposed a learning model to learn more discriminative features in a pair or triplet units.  
%
%
Figure~\ref{fig:preid_dl_methods} shows the taxonomy of the types of models being used for PReID that will be discussed in the coming subsections of this paper. 
\subsection{Single Feature-Learning Based Methods}
\label{subsec:single_feature_learning}
\begin{figure}[!ht]
\centering
\includegraphics[width=0.45\textwidth]{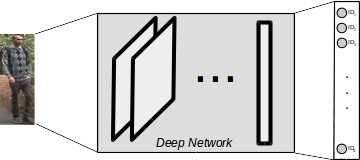}
\caption{Single feature-learning model in PReID system: The model takes the raw image of an individual as input, and computes the probability of the corresponding class of the individual.}
  \label{fig:single_feauture_model}
\end{figure}

A model based on a single feature-learning model or single deep model can be developed similarly to other multi-class classification problems.  In a PReID system, a classification model is designed to determine the probability of identity of an individual that it belongs to~\cite{cheng2018hybrid}. 
Figure~\ref{fig:single_feauture_model} shows an example of a DL based model  for a single feature-learning PReID model. This single stream deep model can be further divided in following categories as being shown in Figure \ref{fig:preid_dl_methods}.
\par
 \textbf{Deep model features fusion with hand-crafted features:} 
 There are number of papers published to boost the performance of PReID by generating deep features. Among them some works additionally involved the hand-crafted features as the complementary features to be fused alongside DL features. These features are further reduced by using traditional dimensionality reduction techniques -- e.g., Principal component analysis(PCA).  
 \par
Wu et al.~\cite{wu2016enhanced} proposed a feature fusion DNN to regularize CNN features, with joint of hand-crafted features. The network takes a single image of size $224 \times 224 \times 3$ pixels as input of the network, and hand-crafted features extracted using one of the state-of-the-art PReID descriptor (best performance obtained from ensemble of local features (ELF) descriptor~\cite{ma2012local}). Then, both extracted features are followed by a buffer layer and a fully-connected layer, which both layers act as a fusion layer. The buffer layer is used for the fusion, which is essential since it bridges the gap between two features with different domains (i.e., hand-crafted features and deep features). A softmax loss layer then takes the output vector of the fully-connected layer to minimizing the cross-entropy loss, and outputs the deep feature representation. The whole network is trained by applying mini-batch stochastic gradient descent algorithm for back propagation.
%
In~\cite{wu2016deep}, two low-level descriptors, SIFT and color-histograms, are extracted from the LAB color space over a set of 14 overlapping patches in size of $32\times 32$ pixels with 16 pixels of stride. Then, a dimensionality reduction method such as PCA, is applied to  scale-invariant feature transform (SIFT)
and color-histogram features to reduce the dimensionality of feature space. Those features are further embedded to produce feature representations using Fisher vector encoding, which are linear separable. One Fisher vector is computed on the SIFT and another one on the color histogram features, and finally, two fisher vectors are concatenated as a single feature vector. A hybrid network builds fully-connected layers on the input of Fisher vectors and employs the linear discriminative analysis (LDA) as an objective function in order to maximize margin between two classes.
\par
A structured graph Laplacian algorithm was utilized in a CNN-based model~\cite{cheng2017deep}. Different from traditional contrastive and triplet loss in terms of joint learning, the structured graph Laplacian algorithm is additionally embedded at the top of the network. They, indeed, formulate the triplet network into a single feature-learning method, and further, used the generated deep features for joint learning on the training sample. Softmax function is used to maximize the inter-class variations of different individual, while the structured graph Laplacian algorithm is employed to minimize the intra-class variations. As the authors pointed out, the designed network needs no additional network branch, which makes the training process more efficient.
Later on, the same authors proposed a structured graph Laplacian embedding approach~\cite{cheng2018deep}; where joint CNNs are leveraged by reformulating structured Euclidean distance relationships into the graph Laplacian form. A triplet embedding method was proposed to generate high-level features by taking into account of inter-personal dispersion and intra-personal compactness. \par

\textbf{Part-based \& Body-based features: } Some works have attempted to generate more discriminant features for their model by extracting features from specific body parts as well as extracting features from whole person's body, that can be used as part of feature vector by fusing it with the deep learning model resultant features. 
In~\cite{su2017pose}, a deep-convolutional model was proposed to handle misalignments and pose variations of pedestrian images. The overall multi-class person re-identification network is composed by two sub-networks: first a convolutional model is adopted to learn global features from the original images; then a part-based network is used to learn local features from an image, which includes six different parts of pedestrian bodies. Finally, both sub-networks are combined in a fusion layer as the output of the network, with shared weight parameters during training. The output of the network is further used as an image signature to evaluate the performance of their person re-identification approach with Euclidean distance. The proposed deep architecture explicitly enables to learn effective feature representations on the person's body part and adaptive similarity measurements.
Li et al.~\cite{li2017learning} designed a multi-scale context aware network to learn powerful features throughout the body and different body parts, which can capture knowledge of the local context by stacking convolutions of multiple scales in each layer. 
In addition, instead of using predefined rigid parts, the proposed model learns and locates deformable pedestrian parts through networks of spatial transformers with new spatial restrictions. Because of variations and background clutter that creates some difficulties in representations based on body-parts, the learning processes of full-body representation is integrated with body-parts for multi-class identification. Chen et al.~\cite{chen2018person} proposed a Deep Pyramidal Feature Learning (DPFL) CNN architecture for explicitly learning multi-scale deep features from a single input image. In addition, a fusion branch over $m$ scales was devised for learning complementary combination of multi-scale features. 
\par
\textbf{Embedding Learning:} Embedding- and attribute-learning approaches have also been considered as a complementary feature by some researchers, where the authors proposed to design a model that can jointly learn additional mid-level features obtained by joint learning of high- and low-level features. 
In~\cite{chen2018improving}, a matching strategy is proposed to compute the similarity between features maps of an individual and corresponding embedding text. Their method is learned by optimizing the global and local association between local visual and linguistic features, where it computes attention weights for each sample. The attention weights are further used by long short-term memory (LSTM) network to enrich the final prediction. It shows that learning based on visual information could be more robust. Similarly, Chi et al.~\cite{su2018multi} proposed a multi-task learning model that learns from embedded attributes. The attribute embedding is employed as a low-rank attribute embedding integrated with low- and mid-level features to describe the person's appearance. On the other hand, deep features are obtained by utilizing a DL framework as a high-level feature extractor. All the features are then learned simultaneously by making use of finding a significant correlation among tasks.
\par
\textbf{Attribute-based Learner: }
A joint DL network is proposed in~\cite{sun2018unified}, which consists of two branches of DL frameworks; in the first branch, the network aims to learn the identity information from person's appearance under a triplet Siamese network (see section~\ref{subsec:triplet_models} for more details), meanwhile, an attribute-based classification is utilized in the second branch to learn a hierarchical loss-guided structure to extract meaningful features. The obtained feature vectors of both branches are then concatenated into a single feature vector. Finally, the person's images in gallery set are ranked according to their feature distances to the final representations.
A method of attention mask-based feature learning is proposed in~\cite{ding2019feature}; the authors proposed a CNN-based hybrid architecture that enables the network to focus on more discriminative parts from person's image. A multi-tasking based solution where the model predicts the attention mask from an input image, and further imposes it on the low-level features in order to re-weighting local features in the feature space.  
\par
\textbf{Semi- and un-supervised learning:} There are also few works related to semi- and un-supervised learning methods that attempted to predict person's identity (i.e., probability of corresponding class label for an individual).  
Li et al.~\cite{li2018unsupervised} proposed a novel unsupervised learning method attempts to replace the fact of manually labelling of data. The method jointly optimizes unlabelled person data within-camera view jointly with cross-camera view under the strategy of end-to-end classification problem. It utilizes deep features generated by a CNN model for the input of their unsupervised learning model.
Wang et al.~\cite{wang2018transferable} proposed a heterogeneous multi-task model by domain transfer learning and addressed the scalable unsupervised learning for the PReID problem. Two branches of CNNs were employed to capture and learn identity and attribute from a person's image simultaneously. The output from both branches are fused with another branch which composed by a shallow NN for a joint learning manner. The information from both branches are inferred into a single attribute space. It showed promising results when their model was trained on a source data set and tested on an unlabeled target data set. 
\par
The approach in ~\cite{xu2018attention} addressed issues such as misalignment and occlusion in PReID. It aims to extract features from different pre-defined person's body-parts, and considers them as pose features and attention aware feature.  Yu et al.~\cite{yu2018unsupervised} proposed a novel unsupervised loss function, in which the model can learn the asymmetric
metric and further embeds it into an end-to-end deep feature learning network.
Moreover, Huang et al.~\cite{huang2018multi} addressed the issue of lack of training data by introducing a multi-pseudo regularized label. The proposed method attempts to generate images based on an adversarial ML techniques, where corresponding class labels are estimated based on semi-supervised learning on a small training set. This could be one possible way of creating synthetic data to train recent deeper NN models.
\par
\textbf{Data Driven: }To address the lack of training data samples, data driven techniques have also been considered for the task of PReID. Xiao et al.~\cite{xiao2016learning} proposed learning deep feature representations from multiple data sets by using CNNs to discover effective neurons for each training set.
They first produced a strong baseline model that works on multiple data sets simultaneously by combining the data and labels from several re-id data sets and training the CNN with a softmax loss. Next, for each data set, they performed the forward pass on all its samples and compute for each neuron its average impact on the objective function. Then, they replaced the standard dropout with the deterministic 'domain guided dropout' to learn generalization by dropping certain neurons during training, and continue to train the CNN model for several epochs. Some neurons are effective only for specific datasets, which might be useless for others due to dataset biases. For instance, the i-LIDS is the only dataset that contains pedestrians with luggage, thus the neurons that capture luggage features will be useless to recognize people from another data set.  
Another technique to overcome the lack of training data samples, data augmentation techniques have proposed. Those techniques are included the methods for flipping, rotating, sheering, etc. which can be applied on original image. Despite those techniques, in~\cite{zhong2019camstyle} a novel data augmentation technique was proposed for PReID, in which a camera style model was developed to generate training data samples via style transfer-learning. 
\subsection{Multi-Stream Network Structure: Pairwise and Triplet Feature-Learning Methods}
\label{subsec:pairwise_feature_learning}
DL models in the PReID problem are still suffering from the lack of training data samples; this is because some of the PReID data sets provide only a few images per individual (e.g., VIPeR dataset~\cite{gray2008viewpoint} which only contains pair of images per person) that makes the model to fail on evaluating the performance of model caused by overfitting problem. Therefore, Siamese networks have been developed to this aim~\cite{li2014deepreid}. 
\par
Siamese network models have been widely employed in PReID due to the lack of training instances in this research area. Siamese neural network (SNN) is a type of NN architectures that contains two or more identical sub-networks (i.e., identical refers to sub-networks when they share the same network architecture, parameters, and weights --a.k.a.~\emph{shared weight parameters}). A Siamese network can be employed as a pairwise model (when two sub-networks are included e.g. \cite{chung2017two, iodice2016strict}), or triplet model (when three sub-networks are present \cite{cheng2016person,liu2016multi}). The output of a Siamese model is a similarity score, which takes place at the top of the network. For instance, a model based on pairwise feature-learning takes two images as its input and outputs similarity score between them. Employing such a siamese model could be an excellent solution to train on existing PReID data set\cite{zheng2016person}, when a few training samples are available.
\begin{figure}[!ht]
\centering
\includegraphics[width=0.7\textwidth]{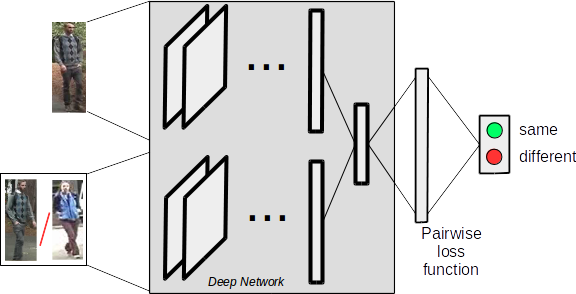}
\caption{Pairwise-loss feature-learning model.}
  \label{fig:pairwise_model}
\end{figure}
These models can be divided in the way we discussed single stream models in previous section \ref{subsec:single_feature_learning} and as shown in Figure \ref{fig:preid_dl_methods}. However, the rest of this section is organized in three subsections. First we gave a brief explanation of the similarity functions engaged in DL-based PReID methods. These are essential concepts to compute the distance of similarity between the output of the multi (two/three) models from the given multi input images during training DL model. In the second subsection, we described the published DL-based work for the pairwise methods followed by triplet methods in third subsection. Both of these pairwise and triplet follows the single stream feature learning approaches. 
\subsubsection{Similarity functions}
\label{subsec:similarity_funcs}
In order to measure the similarity between the pair of input images within a siamese network, typically, an objective function is utilized. 
An objective function (a.k.a.~\textit{loss} or~\textit{cost function}) aims to map intuitively some values into a one single real number. This represents a cost which is associated with those values. Techniques like NNs seek to minimize a loss function optimally.
When a loss function is used for a siamese model, it depends on the type of the utilized model (i.e., either a pairwise or triplet model).
%
\par
In the case of a pairwise model,
let $X=\{ x_1, x_2, \dots, x_n \}$ and $Y=\{y_1, y_2, \dots, y_n \}$ be a set of images and corresponding labels for each individual, which can be formulated as 
\begin{equation}
\label{eq:pairwise_pos_neg}
I(x_i,x_j)=  \bigg\{ 
\begin{array}{lll}
positive & if & y_i=y_j,  \\
negative & if  & y_i!=y_j
\end{array}
\end{equation}
the goal is to minimize the relative distance between the matched pairs and maximize with the mismatched pairs,for given a pair of image representations, $x_1$ and $x_2$, and corresponding labels $y \in \{+1,-1\}$.

Among existing loss functions for pairwise classification models, Hinge, and Contrastive loss functions have widely utilized in this vein. 
%
Hinge loss function refers to maximum-margin classification; the output of this loss is become zero when the distance similarity of the positive pairs is greater than the distance of the negative ones with respect to the margin $m$. This loss is defined as follow
\begin{equation}
\label{eq:pairwise_hinge_loss}
I(x_1,x_2,y)=  \bigg\{ 
\begin{array}{lll}
 \left\lVert x_1-x_2 \right\rVert & if & y=1  \\
max(0,m-  \left\lVert x_1-x_2 \right\rVert ) & if & y=-1
\end{array}
\end{equation}
The Cosine similarity loss function maximizes the cosine value for positive pairs and reduce the angle in between them, and at the same time, minimize the cosine value for the negative pairs when the value is less than margin. 
\begin{equation}
\label{eq:pairwise_cosine_loss}
I(x_1,x_2,y)=  \bigg\{ 
\begin{array}{lll}
max(0,cos(x_1,x_2)-m) & if & y=1 \\
1-cos(x_1,x_2) & if & y=-1
\end{array}
\end{equation}
A Contrastive loss function minimizes meaningful mapping from high to low dimensional space maps by keeping the similarity of input vectors of nearby points on its output manifold and dissimilar vectors to distant points. Accordingly, the loss can be computed as:
\begin{equation}
\label{eq:pairwise_Contrastive_total_loss}
I(x_1,x_2,y)= (1-y)\frac{1}{2}(D)^2 + (y)\frac{1}{2}\{max(0,m-D)\}^2
\end{equation}
where $m>0$ is a margin parameter acting as a boundary, and $D$ is a distance between two feature vector that is computed as $D(x_1,x_2)= \left\lVert x_1-x_2 \right\rVert_2$.
In order to compute the average of total loss for each above-mentioned pairwise loss functions, it can be computed as
\begin{equation}
\label{eq:pairwise_cosine_total_loss}
\mathcal{L}(X_1,X_2,Y)= -\frac{1}{n}\sum_{i=1}^n I(x_i^1,x_i^2,y_i)
\end{equation}
\\
For a triplet model, an objective function is used to train the network models, which creates a margin between the distance metric of positive pair and distance metric of negative pair. For this type of Siamese model, a softmax layer is employed at the top of the network on both distance outputs. 
Let $O_i=\{ (I_i, I^+_i, I^-_i)\}^N_{i=1}$ be a set of triplet images, in which $I_i$ and $I^+_i$ are images of the same person, and $I^-_i$ is a different person. 
A triplet loss function is used to train the network models, which makes the distance between $I_i$ and $I^+_i$less than the mismatched pairs $I_i$ and $I^-_i$ in the learning feature space.
In the triple-based models, Euclidean loss function is commonly used as a distance metric function. The loss function under \emph{L2} distance metric, and denoted as $d(W,O_i)$; where $W={W_i}$ is the network parameters, and $F_w(I)$ represents the network output of image $I$. The difference in the distance is computed between the matched pair and the mismatched pair of a single triplet unit $O_i$:
\begin{equation}
\label{eq:loss_for_1}
d(W,I_i)= \lVert{F_W(I_i)-F_W(I_i^+)}\rVert^2 -  \lVert{F_W(I_i)-F_W(I_i^-)}\rVert^2
\end{equation}
Moreover, the Hinge loss function is another widely used distance measurements. This loss  function is a convex approximation in range of 0-1 ranking error loss, which approximate the model's violation of the ranking order specified in the triplet.
\begin{equation}
\label{eq:loss_for_hinge}
\mathcal{L}(I_i,I^+_i,I^-_i)= max(0,g+D(I_i,I^+_i)-D(I_i,I^-_i))
\end{equation}
where $g$ is a margin parameter that regularizes the margin between the distance of the two image pairs: $(I_i , I_i^+ )$ and $(I_i , I_i^- )$, and $D$ is the euclidean distance between the two euclidean points.
\subsubsection{Pairwise-loss methods}
\label{subsec:pairwise_models}
Several works rely upon pairwise modeling in order to learn features from small set of training data. To this aim, some works proposed novel deep learning architectures for learning in pairwise manner. This types of learning treats PReID task as a binary-class classification problem~\cite{li2014deepreid,ahmed2015improved,wu2016personnet}. 
In~\cite{zhang2014people}, a Siamese pair-based model takes two images as the input of two sub-networks, where two networks are locally connected to the first convolutional layer. They employed a linear SVM on the top of the network instead of using a softmax activation function to measure the similarity of input images pair as the output of the network. \par
In~\cite{yi2014deep}, a siamese neural network (SNN) has been designed to learn pairwise similarity. Each input image of a pair was first partitioned into three overlapping horizontal parts. The part pairs are matched through three independent Siamese networks and finally, are combined at score level.
Li et al.~\cite{li2014deepreid} proposed a deep filter pairing NN to encode photometric transformation across camera views. A patch matching layer is further added to the network to multiple convolution feature maps of pair of images in different horizontal stripes. 
Later, Ahmed et al.~\cite{ahmed2015improved} improved the pair-based Siamese model in which the network takes pairs of images as the input, and outputs the probability of whether two images in the pair are referred to the same or different persons. The generated feature maps are passed through a max-pooling kernel to another convolution layer followed by a max-pooling layer in order to decrease the size of the feature map. Then a cross-input neighborhood layer computes the differences of the features in neighboring locations of the other image. 
\par
\textbf{New Architectures:} Wang et al.~\cite{wang2016joint} developed a CNN model to jointly learn single-image representation (SIR) and cross-image representation (CIR) for PReID. Their methodology relied on investigating two separate models for comparing pairwise and triplet images (explained in next section) with similar deep structure. Each of these models configured with different sub-networks for SIR and CIR learning, and another sub-network shared by both SIR and CIR learning. For the pairwise comparison, they used the Euclidean distance as loss function to
learn SIR and formulated the CIR learning that provides a binary classification problem and employs the standard SVM to learn CIR as its loss function. It uses the combination of both loss functions as the overall loss function of pairwise comparison. For the triplet comparison, the loss function to learn SIR makes the distance between the matched pairs lower than the mismatched pairs. The CIR learning formulates a learning-to-rank problem and employs the 'RankSVM' as its loss function. To this end, the combination of both learning methods is used as the overall loss function of the triple comparison. The shared sub-networks share parameters during the training stage.
\par
Wang et al.~\cite{wang2016person} proposed a pairwise Siamese model by embedding a metric learning method at the top of the network to learn spatiotemporal features. The network takes a pair of images in order to obtain CNN features and outputs whether two images belong to the same person by employing the quadratic discriminant analysis method. 
\par
To handle the multi-view person images and learning in pairwise manner, a new deep multi-view feature learning (DMVFL) model was proposed in~\cite{tao2017deep} to combine handcrafted features (e.g., local maximal occurrence (LOMO)~\cite{liao2015person}) with deep features generated by CNN-based model; and embedding a metric distance method at the top of the network in order to learn metric distance. To this aim, the cross-view quadratic discriminant analysis (XQDA) metric learning method~\cite{liao2015person} is utilized to jointly learn handcrafted and DL features. In this manner, it is possible to investigate how handcrafted features could be influenced by deep CNN features. Further, 
a two-channel CNN with a new component named Pyramid Person Matching Network (PPMN) is proposed ~\cite{mao2017pyramid} with the same architecture of GoogLeNet. The network takes a pair of images and extracts semantic features by convolutional layers. Finally, the Pyramid Matching Module learns the corresponding similarity between semantic features based on multi-scale convolutional layers. 
A Strict Pyramidal Deep Metric Learning approach was proposed in ~\cite{iodice2016strict} in which a Siamese network is composed by two strict pyramidal CNN blocks with shared parameters between each and produced salient features of an individual as the output of the network. The objective was to present a simple network structure that can perform well despite fewer parameters and having a trade-off between lower computational and memory costs concerning the other NNs.
In~\cite{shen2018deep}, a Siamese pairwise model is designed as a manner of re-ranking approach where a CNN model was adopted to generate high-level features. The obtained feature maps from both sub-networks are mapped into a single feature vector. It is then divided into $K$ feature groups, in which the output of the network is equivalent to the number of feature groups and presented as the similarity scores corresponded to each pair.
Shen et al.~\cite{shen2018end} addressed the issue of spatial information from a person's appearance. They utilized Kronecker-product matching (KPM) for aligning feature maps of each individual and further used them in order to generate matching confidence maps between pairs of images. At the beginning of their model, two CNN models are utilized to generate feature maps for each pair of images, separately. Then, the obtained feature maps are used in the KPM method to generate wrapped feature maps. The difference between two feature vectors is then mapped between generated feature maps from the first image and wrapped feature maps by using simple element-wise subtraction. Also, a self-residual attention learning is employed on the feature maps of the first image. Finally, the computed features maps are further mapped into a single feature map by using element-wise addition. The final feature map is then followed by an element-wise square, batch normalization, and a softmax layer yields the final probability score between a pair of images.
\par
A pairwise multi-task DL based model is proposed ~\cite{mclaughlin2016person} to use a separate softmax for each auxiliary task as identification, pose labeling, and each attribute labeling task. A CNN is used to generate image representations in which a single image of size $64 \times 64$ is used as the input. It takes a pair of images by embedding a specific cost function for each. For instance, they used a softmax regression cost function for each task in which a multi-class linear classifier calculates the probability of a person's identity. They minimized the cost function using SGD concerning the weights of each task; then, the linear combination overall cost functions is presented as the final cost function of the network. The designed network is composed of three convolutional layers, and two max-pooling layers, followed by a final fully-connected layer as the output of the network. The hyperbolic tangent activation function was used between each convolutional layer, while a linear layer was used between the final convolutional layer and the fully-connected layer. The activation of neurons in the fully-connected layer gives the feature representation of the input image; dropout regularization was used between the final convolutional layer and the fully-connected layer. 
\par
A novel Siamese Long-Short Term Memory (LSTM) based architecture was proposed in~\cite{varior2016siamese} which aims to leverage contextual dependencies by selecting relevant contexts to enhance discriminative capabilities of the local features. They proposed a pairwise Siamese model, which contains six LSTM models for each sub-network. First, each image is divided into six horizontal non-overlapping parts. From each part, an image representation is extracted by using two state-of-the-art descriptors (i.e., LOMO and Color Names). Each feature vector is separately fed to a single LSTM network with the share parameters. The outputs from each LSTM network are combined, and the relative distance of subnets is computed by the contrastive loss function. The whole pairwise network is trained with the mini-batch SGD algorithm.
\par
\textbf{Part- and Body-Based Features Fusion:} Furthermore, some researches have also considered multi-scale and multi-part feature learning from person's images.
Wang et al.~\cite{wang2016deeplist} designed an ensemble of multi-scale and multi-part with CNNs to jointly learn image representations and similarity measure. The
The network takes two person images as the input of the network and derived the full scale, half scale, top part, and middle part image pairs from the original images. The network outputs the similarity score of the pair images. This architecture is composed of four separate sub-CNNs with each sub-CNN embedding images of different scales or different parts. The first sub-CNN takes full images of size $200 \times 100$ and the second sub-CNN takes down-sampled images of size $100 \times 50$. The next two sub-CNNs take the top part and middle part as input, respectively. Four sub-CNNs are all composed of two convolutional layers, two max-pooling layers, a fully-connected layer, and one L2-normalization layer. They obtained the image representation from each sub-CNN and then calculated their similarity score. The final score is calculated by averaging four separate scores; A ReLU activation function is used as the neuron activation function for each layer, dropout layer is used in the fully-connected layer to reduce the risk of an over-fitting problem.
\par
Liu et al.~\cite{liu2016end} utilized a deep model to integrate a soft attention-based model into a Siamese network. The model focuses on local parts of input images on the pair-based Siamese model. 
\par
\textbf{Multi-Scale Learning Models: } A multi-scale learning model was proposed in  
~\cite{qian2017multi} in which the proposed approach can learn discriminant feature from multi-scales of an image in different levels of resolution. A saliency-based learning strategy is adopted to learn important weight scales. 
In parallel with the pairwise Siamese model, which aims to distinguish whether a pair of images belong to the same person or not, a tied layer is also used between each layer of each branch in order to verify the identity of an individual. 
The designed model consists of five components: tied convolutional layers, multi-scale stream layers, saliency-based learning fusion layer, verification subnet, and classification sub-network. The same authors, lately, proposed another approach in~\cite{qian2019leader} in order to learn pedestrian features from different resolution levels of filters over multiple locations and spatial scales.
\par

A patch-based feature learning method was proposed in~\cite{shi2016embedding}. A pairwise Siamese network takes a CNN features pair as input and outputs the similarity value between them by applying the cosine and Euclidean distance function. Each sub-network contains a CNN-based model to obtain deep features of each input image pair, and then, each image is split into three overlapping color patches. The deep network built in three different branches, and each branch takes a single patch as its input; finally, the three branches are concluded by a fully-connected layer. 
\par
Some works attempted to adopt metric- and transfer-learning methods in pairwise feature-learning models. 
Chen et al.~\cite{chen2016deep} proposed a deep ranking model to jointly learn image representation and similarities for comparing pairwise images. To this aim, a deep CNN is trained to assign a higher similarity score to the positive pair than any negative pairs in each ranking unit by utilizing the logistic activation function, which is employed as $\sigma(x)=log_2(1+2^{-x})$.
They first stitched a pair of images horizontally to form an image which is used as an input, and then, the network returns a similarity score as its output. 
A Deep Hybrid Similarity Learning model (DHSL)~\cite{zhu2017deep} based on CNN is proposed to learn the similarity between pair of images. This two-channel CNN with ten layers aims to learn pair feature vectors, discriminate input pairs to minimize the network's output value for similar pair images, and to maximize for different ones. A new hybrid distance method using element-wise absolute difference and multiplication is proposed to improve the CNN in similarity metrics learning. 
\par
\textbf{Transfer learning:} It is a technique which consists of fine-tuning the parameters
of a network that has been already trained on a different dataset, in order to adapt it into a new system.
Franco et al.~\cite{franco2016coarse} proposed a coarse-to-fine approach to achieve generic-to-specific knowledge through transfer learning. The approach follows three steps: first a hybrid network is trained to recognize a person, then another hybrid network employed to discriminate the gender of person; the output of two networks are passed through the coarse-to-fine transfer learning method to a pairwise Siamese network to accomplish the final PReID goal in terms of measuring the similarity between those two features.
Later, the same authors proposed a different type of features based on convolutional covariance descriptor (CCF) ~\cite{franco2017convolutional}. They intend to obtain a set of local covariance matrices over the feature maps extracted by the hybrid network under the same strategy of the above-proposed method.

\subsubsection{Triplet-loss methods}
\label{subsec:triplet_models}
\begin{figure}[!ht]
\centering
\includegraphics[width=0.45\textwidth]{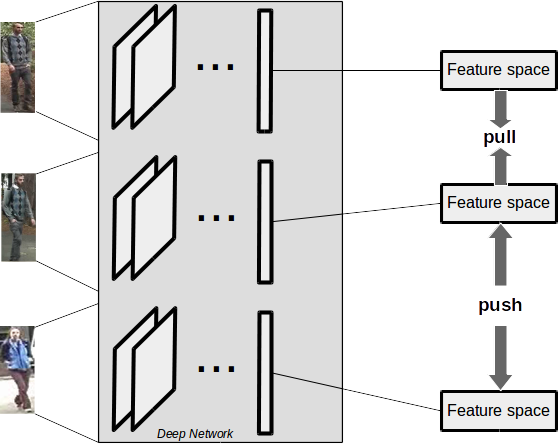}
\caption{Triplet-loss feature-learning model.}
  \label{fig:triplet_model}
\end{figure}

Several works proposed novel PReID systems based on deep learning architecture for learning in a triplet manner. Triplet models mainly introduced for image-retrieval~\cite{wang2014learning} and face recognition~\cite{schroff2015facenet} problems. Such that model takes three images of individual, in a formation as a triplet unit, aiming to minimize the relevant similarity distance between the same person, and maximize from different one. Figure~\ref{fig:triplet_model} shows a basic triplet model. 
This type of models either can share weights or keep them independent. 

\par
Ding et al.~\cite{ding2015deep} is the first work in PReID task which adopted a triplet deep CNN-based models to produce robust feature representations from raw images. It takes input image size of $250 \times 100$ pixels as a triplet unit, where the weights are shared  between each sub-networks. It aims to maximize the relative distance between pairs of images of the same person and a different person under $L_2$ loss function. The model is trained with the SGD algorithm with respect to the output feature of the network.
\par
A learning approach was proposed in~\cite{zhang2015bit} to reformulate a multi-tasking problem in PReID task; whereby the method considered as a joint system overall image retrieval technique across disjoint camera views jointly with deep features and hash-learning functions. A deep architecture of NN was utilized to produce the hashing codes with the weight matrix by taking raw images of size $250 \times 100$ pixels as input of the network. The network was trained in a triplet manner for similarity feature-learning to enforce that the images of same person should have similar hash codes. For each triplet unit, it maximize the margin between the matched pairs and the mismatched pairs. It uses Alexnet pre-trained network that consists of ten layers: the first six layers form the convolution-pooling network with rectified linear activation and average pooling operation. They used 32, 64, and 128 kernels with size $5 \times 5$ in the first, second, and third convolutional layers and the stride of 2 pixels in every convolution layer. The stride for pooling is 1 and they set the pooling operator size as $2 \times 2$. The last four layers consists of two fully-connected layers, and a tangent like layer to generate the output as the hash codes, and an element-wise connected layer to manipulate the hash code length by weighting each bin of the hashing codes. The number of units set 512 in the first fully-connected layer and the output of the second fully-connected layer equals to the length of hash code. The activation function of the second fully-connected layer is the tanh-like function, while
ReLu activation function is adopted for the others.

\par
\textbf{Part- and Body-based features: } Cheng et al.~\cite{cheng2016person} proposed a triplet loss function in which the network takes a triplet unit of images as input, and jointly learn from the global full-body and local body-parts features as robust representation. The fusion of these two types of features at the top of the network is presented as the output of the network. The utilized CNN model begins with a convolution layer, divided into four equal parts, and each part forms the first layer of an independent body-part channel aiming to learn features from that body part. The four body-part channels with the full-body channel constitute five independent channels that are trained separately from each other (with no parameter sharing between each). At the top of the network, the outputs obtained from the five separate channels are concatenated into a single vector, and is passed through a final fully-connected layer. 
Bai et al.~\cite{bai2017deep} proposed a deep-person model to generate global- and part-based feature representations of a person's body. Each image of triplet unit is fed into a backbone CNN to generate low-level features with the shared parameters. Output features of the backbone network are further fed into a two-layer Bidirectional LSTM aiming to generate a part-based feature representation; an LSTM is adopted because of its discriminative ability of part representation with contextual information, handling the misalignment with the sequence-level person representation.
At the same time, layer output features are also fed into another network branch, with a global average pooling, a fully-connected, and a Softmax layer for global feature learning. Finally, output features learn similarity distances under a triplet loss function by adopting another branch of network during the training of the whole network. 
A coherent and conscious DL approach was introduced that is able to cover whole network cameras~\cite{lin2017consistent}. The proposed approach aims to seek the globally optimal matching over different cameras. The deep features are generated over full body and part body under a triplet framework, in which each image within a triple unit is presented with a sample image of one camera view, while the other images are presented from other camera views. Once deep features are generated, the cosine similarity is used to obtain similarity scores between them, and afterward, the gradient descent is adopted to obtain the final optimal association. All calculations are involved in both forward and backward propagation to update CNN features.
\par
\textbf{Attribute based models:} An attribute-based method is proposed by Chen et al.~\cite{chen2018deep} that uses embedding learning to drive attributes and identity annotations from a person's appearance, whereby two embedding-based CNNs are learned, simultaneously. The pre-defined attributes of this work, mainly, rely on pedestrian's appearance in order to extract similar cues between the same person --i.e., if a pedestrian wears a red T-shirt and/or a black backpack at the same time.   An improved triplet loss is used to learn the fusion of them. Due to spatial variations caused by
pose/ view point changes, the proposed model is robust in terms of the diversity on the appearance of pedestrian attributes.
A multi-image re-ranking approach was proposed ~\cite{yuan2019multi} in which an image pool was formed to collect the images of each identity. It uses a CNN-based model in a triplet manner, where the feature vectors obtained from the network during the re-ranking step, is used to compute the similarities between image pools and templates in gallery set. 
\par
\textbf{Multi-scale Learning: } Multi-part and multi-scale approaches are also considered in a triplet manner. Liu et al.~\cite{liu2016multi} proposed a multi-scale triplet network by employing a single CNN-based network and two shallow NNs (i.e., to produce less invariance and low-level appearance features from images), with shared parameters between them. The deep network designed with five convolutional layers, five max-pooling layers, two local normalization layers, and three fully-connection layers, while each shallow network composed by two convolutional layers followed by two pooling layers. 
The output of each network is further combined at an embedding layer in order to generate final feature representation.
Wu et al.~\cite{wu2019attention} proposed an attention-multi-scale deep learning technique for joint-learning of low- and high-level features. The proposed deep architecture consists of five branches in which the first branch of the network used to learn deep features via attention block. A triplet and four classification losses adopted to learn the global descriptor through the second and third branches, respectively. Furthermore, a multi-scale feature learning is applied in the fourth and fifth branches of their network. 
\par
\textbf{Semi-supervised Approach:} A novel semi-supervised Deep Attribute Learning approach is proposed in ~\cite{su2016deep}, which contains three deep CNN-based networks and the whole network is trained with attributes triplet loss. The first network is trained on an independent data set to predict the predefined attributes (e.g.,). Second network is trained on another data set plus the predicted attributes labels from the first sub-network. Finally, the last network is used to distinguished attributes and trained on another data sets with individual class labels. The proposed method is more reliable on real-world scenario for PReID, in which the proposed solution can be performed onto another unknown target dataset. 

\section{Results and Open Issues}
\label{sec:experimental_comparison}
\begin{table*}[!ht]
\caption{Comparison of existing DL models based on Rank-1 recognition rates PReID. Type of models (single, pairwise, and triplet) are denoted by $S$, $P$, and $T$ and colored by \textcolor{blue}{blue},~\textcolor{red}{red}, and \textcolor{green}{green}, respectively. This table is best viewed in color.}
\label{tab:cmc_firstRank}
\centering
\tiny{
\begin{tabular}{|r|c|c||c|c|c|c|c|c|c|c|} \hline
\multicolumn{3}{|c||}{} & \multicolumn{8}{c|}{\textbf{Rank-1 Recognition Rate on specific Datasets}} \\ \hline

\textbf{Ref.\#}  & \textbf{Year}  & \textbf{Model} & \textbf{VIPeR} & \textbf{CUHK01}    &  \textbf{CUHK03}   & \textbf{i-LIDS} & \textbf{PRID-2011}  & \textbf{CAVIAR} & \textbf{MARS} & \textbf{Market-1501} \\ \hline
Li~\cite{li2014deepreid}          & 2014      &$\textcolor{red}{P}$ &       --         & 20.65   & --      & --      & -- & -- &--&-- \\
Zhang~\cite{zhang2014people} & 2014    &$\textcolor{red}{P}$&                      12.50    & -- & --  & --  & --  & 7.20 &-- & --\\  

Yi~\cite{yi2014deep} & 2014& $\textcolor{red}{P}$ &   -- & 28.23 & --  & --  & --  & --  &-- &--   \\ 
\hline 

Ahmed~\cite{ahmed2015improved}       & 2015 &$\textcolor{red}{P}$ &    34.81 &   65.00 & 54.74& --  &  --    &  -- &--&-- \\     
Ding~\cite{ding2015deep}& 2015 &$\textcolor{green}{T}$  & 40.50  &  -- &  -- & 52.10 &  -- & -- &-- &--  \\ 

Zhang~\cite{zhang2015bit}        & 2015   &    $\textcolor{green}{T}$ &         --  & --  & 18.74   &  -- & --  & -- &--&-- \\
Shi~\cite{shi2015constrained}& 2015 &$\textcolor{red}{P}$  & 40.91  &  86.59 &  59.05 & -- &  -- & -- &-- &--  \\ \hline

Liu~\cite{liu2016end}   & 2016  &$\textcolor{red}{P}$&            --              & 81.40    & 65.65  &  --  &  --  &  -- &--&48.24\\ 
Cheng~\cite{cheng2016person}& 2016 &   $\textcolor{green}{T}$     &   \textcolor{green}{47.80}      & 53.70    &  --    &  --   & 22.00  & -- &--&-- \\ 

Chen~\cite{chen2016deep}     & 2016   &$\textcolor{red}{P}$   &               \textcolor{red}{52.85}     & 57.28   &  --   &  --   &  -- &   \textcolor{red}{\textbf{53.60}} & --& --\\ 

Wu~\cite{wu2016enhanced}   & 2016  &   $\textcolor{blue}{S}$&      51.06            & 55.51   &  --   &  --   &  66.62 & -- &--&--\\ 

Xiao~\cite{xiao2016learning}     & 2016   &$\textcolor{blue}{S}$  &          38.60          & 66.60   & 75.30   &  64.60 & 64.00 & -- &--&--\\ 
Wu~\cite{wu2016personnet} & 2016   &$\textcolor{red}{P}$    &        --          & 71.14   & 64.90  &  -- & -- & -- &--& 37.21 \\
Li~\cite{li2016discriminative}    & 2016     &$\textcolor{red}{P}$&             --       & --   & --   & -- & --  & -- & --& 59.56 \\ 

Shi~\cite{shi2016embedding}    & 2016     &$\textcolor{red}{P}$&            40.91        & 69.00   & --   & -- & --  & -- & --& --\\ 

Varior~\cite{varior2016siamese} & 2016 &$\textcolor{red}{P}$& 42.40 & -- & 57.30 & --  & --  &   -- &--&61.60 \\ 

Wang~\cite{wang2016deeplist}   & 2016        &  $\textcolor{red}{P}$&    40.51       & 57.02   & 55.89  &  --   &  -- & -- &-- &-- \\ 
Wang~\cite{wang2016joint}         & 2016     &      $\textcolor{red}{P}$ &   29.75   & 58.93   & 43.36   & --  & --  & -- & --& --\\ 
Wang~\cite{wang2016joint}         & 2016     &       $\textcolor{green}{T}$   & 35.13   & 65.21  &    51.33 & --  & --  & -- & --& --\\ 
Franco~\cite{franco2016coarse}     & 2016      &$\textcolor{red}{P}$&     --     &      44.94   & 63.51  & 62.30  & 53.33 &  -- &--&--\\ 
Wu~\cite{wu2016deep}          & 2016       &    $\textcolor{blue}{S}$   &  44.11  & 67.12  &  --   &  --  &  -- & -- & --& 48.15 \\ 
Wang~\cite{wang2016person}          & 2016  &$\textcolor{red}{P}$&         --       & 38.28  & 27.92  & -- & --  & -- &-- &-- \\ 
Su~\cite{su2016deep}             & 2016          &$\textcolor{green}{T}$&       43.50   & --  & --  & --  & 22.60 &  -- &--&--\\ 

Mclaughlin~\cite{mclaughlin2016person}  & 2016 &         $\textcolor{red}{P}$ & 33.60  & -- &  -- & -- &  -- &   -- & --& --\\ 
McLaughlin~\cite{mclaughlin2016recurrent}&2016&$\textcolor{red}{P}$& --   & -- & --& \textcolor{red}{85.00}  & \textcolor{red}{70.00}  &   -- &--&--\\ 
Liu~\cite{liu2016multi}&2016&$\textcolor{green}{T}$&  --   & -- & --& -- & -- &   -- &--&55.40\\ 
Iodice~\cite{iodice2016strict}&2016&$\textcolor{red}{P}$&  18.04   & -- & --& -- & -- &   -- &--&--\\
\hline

Su~\cite{su2017pose}&2017&$\textcolor{blue}{S}$&   51.27   & -- & 78.29& --  & --  &   -- &--& 63.14\\ 
Li~\cite{li2017learning}& 2017 &$\textcolor{blue}{S}$&  38.08   & -- & 74.21 & --  & --  &   -- &71.77&80.31\\ 
Franco~\cite{franco2017convolutional}    & 2017    &$\textcolor{red}{P}$&  --            & 63.85  & 63.90  & 55.85 &  -- & -- &-- &-- \\ 
Qian~\cite{qian2017multi} & 2017 &$\textcolor{red}{P} \& \textcolor{blue}{S}$& 43.30    & 79.01 & 76.87 & 41.00 & 65.00 & -- &-- &--  \\ 
Zhu~\cite{zhu2017deep} & 2017 &$\textcolor{red}{P}$& 44.87    &  -- &  -- &  -- & -- &  --   & --& --\\  
Cheng~\cite{cheng2017deep} & 2017 &$\textcolor{green}{T}$&    -- & 70.09 & 84.70 &  -- &  -- & --     &--&83.6\\  
Tao~\cite{tao2017deep} & 2017 &$\textcolor{red}{P}$&  46.00    &  -- & --  & --  &  -- & --  &&\\  
Mao~\cite{mao2017pyramid} & 2017 &$\textcolor{red}{P}$&   45.82    & \textcolor{red}{\textbf{93.10}} & 85.50 & --  & --  &   -- &--&-- \\ 

Lin~\cite{lin2017consistent} & 2017 &$\textcolor{green}{T}$&  --   & -- & --& --  & --  &   -- &--&81.15\\ 

Bai~\cite{bai2017deep}&2017&$\textcolor{green}{T}$&   -- & -- & \textcolor{green}{91.50} & --  & -- & -- &--&92.31\\ 

Chung~\cite{chung2017two}&2017&$\textcolor{green}{T}$&  --   & -- & --& 60.00  & 78.00  &   -- &--&--\\ 
Chen~\cite{chen2017person}&2017&$\textcolor{blue}{S}$&  50.30   & 74.50 & 84.30 & --  & -- & -- & -- & 68.70 \\
\hline
Li~\cite{li2018unsupervised}&2018&$\textcolor{blue}{S}$&   -- & -- & 44.70 & 26.70  & 49.40 & -- &43.80 &63.70  \\
Chen~\cite{chen2018improving} &2018&$\textcolor{blue}{S}$&   -- & 84.08 & 92.50 & --  & -- & -- &--&93.30
\\ 
Chi~\cite{su2018multi}&2018&$\textcolor{blue}{S}$&   45.40 & -- & -- & 56.40  & 21.00 & -- &--&--\\ 
Sun~\cite{sun2018unified}&2018&$ \textcolor{blue}{S}$&   -- & -- & -- & --  & -- & -- &--&87.05\\ Wang~\cite{wang2018transferable}&2018&$ \textcolor{blue}{S}$&   38.50 & -- & -- & --  & -- & 34.80 &--&58.20\\ 
Xu~\cite{xu2018attention}&2018&$ \textcolor{blue}{S}$&   -- & \textcolor{blue}{88.07} & 91.39 & --  & -- & -- &--&88.69\\
Chen~\cite{chen2018person}&2018&$\textcolor{blue}{S}$&   --& -- & 86.70  & --  & -- & -- &--&88.90\\
Shen~\cite{shen2018deep}&2018&$\textcolor{red}{P}$&   -- & -- & \textcolor{red}{94.90} & --  & -- & -- &--& 82.50 \\ 
Huang~\cite{huang2018multi}&2018&$\textcolor{blue}{S}$&  54.65& 78.83 & 81.28  & --  & -- & -- &--&87.96\\
Shen~\cite{shen2018end}&2018&$\textcolor{red}{P}$&   -- & -- & 93.40 & --  & -- & -- &--&90.10 \\ 
Chen~\cite{chen2018deep}&2018&$\textcolor{green}{T}$&   -- & -- & 65.00 & -- & -- &--& -- & --\\
Cheng~\cite{cheng2018deep}&2018&$\textcolor{green}{T}$&   -- & \textcolor{green}{70.90} & 84.70 & -- & -- &--& --& 83.60 \\
Chen~\cite{chen2018group}&2018&$\textcolor{red}{P}$&   --& -- & 90.20  & --  & -- & -- &--&\textcolor{red}{93.50}\\ 
Li~\cite{li2018harmonious}&2018&$\textcolor{blue}{S}$&   --& -- & 44.40  & --  & -- & -- &--&91.20\\
Yu~\cite{yu2018unsupervised} &   2018    & $\textcolor{blue}{S}$ & 34.15 & 69.00  & 45.82  & -- & -- &--&-- &60.24\\
\hline  
Ding~\cite{ding2019feature} &   2019    & $\textcolor{blue}{S}$ & -- & 42.60  & --  & -- & -- &--&--&86.00\\
Yuan~\cite{yuan2019multi} &   2019    & $\textcolor{green}{T}$ & -- & --  & --  & \textcolor{green}{66.00} & \textcolor{green}{81.00} &--&--&--\\
Xiong~\cite{xiong2019good} &   2019    & $\textcolor{blue}{S}$ & -- & --  & 63.50  & -- & -- &--&--&92.50\\
Zhong~\cite{zhong2019camstyle}&   2019    & $\textcolor{blue}{S}$ & -- & --  & --  & -- & -- &--&--&89.49\\
Yao~\cite{yao2019deep}&   2019    & $\textcolor{blue}{S}$ & \textcolor{blue}{\textbf{56.65}} & --  & 82.75  & -- & -- &--&--&88.02\\
Chen~\cite{chen2019person}&   2019    & $\textcolor{blue}{S}$ & -- & --  & --  & -- & -- &--&--&94.50\\
Zheng~\cite{zheng2019pose}&   2019    & $\textcolor{blue}{S}$ & -- & --  & 45.88  & -- & -- &--&--&87.33\\
Zheng~\cite{zhang2019scan}&   2019    & $\textcolor{blue}{S}$ & -- & --  & --  & \textcolor{blue}{\textbf{88.00}} & \textcolor{blue}{\textbf{95.30}} &--&\textcolor{blue}{\textbf{87.20}} &--\\
Qian~\cite{qian2019leader}&   2019    & $\textcolor{blue}{S}$ & -- & 87.55  & \textcolor{blue}{\textbf{95.84}}  & -- & -- &--&-- &\textcolor{blue}{95.34}\\
Wu~\cite{wu2019attention}&   2019    & $\textcolor{green}{T}$ & -- & --  & 81.00  & -- & -- &--&-- &\textcolor{green}{\textbf{95.50}}\\

\hline
\end{tabular}}
\end{table*}

\textbf{Performance Measure} To evaluate the performance of a PReID system, the cumulative matching characteristic (CMC) curve is typically calculated and demonstrated as a standard recognition rate of which the individuals are correctly identified within a sorted, ranked list. In other words, a CMC curve is defined as the probability that the correct identity is within the first $\mathrm{'r'}$ ranks, where $\mathrm{r}=1, 2, \ldots, n$, and $n$ is the total number of template images involved during the testing of a PReID system.
By definition, the CMC curve increases with $\mathrm{'r'}$, and eventually equals 1 for $\mathrm{r} = n$.
%
\par
\begin{figure}[!b]
\centering
\includegraphics[width=0.45\textwidth]{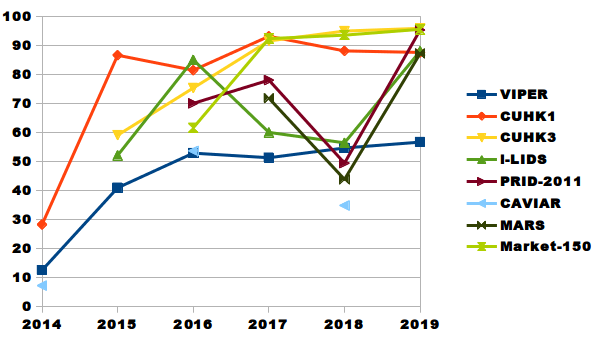}
\caption{Recognition accuracy on the benchmark data sets over the years.}
  \label{fig:ds_accu_year}
\end{figure}

We attempted to collect the original CMC curves presented at each of the previously discussed works for the sake of a comprehensive comparison. However, the CMC curves of most of those works are not available publicly. We therefore listed in table~\ref{tab:cmc_firstRank} and compared only the first-rank (Rank-1) recognition rate of existing deep PReID techniques since 2014 till date. 
Rank-1 has a higher importance in PReID due to the reason that the system needs to recognize the person from the limited hard to recognize available data in the first glance. 
Further, we showed the type of classification models i.e. single, pairwise, and triplet denoted in the table by $S$, $P$, and $T$, and colored by \textcolor{blue}{blue},~\textcolor{red}{red}, and \textcolor{green}{green}, respectively. The global best results among the methods are shown in bold. Moreover, Fig.~\ref{fig:ds_accu_year} demonstrates the Rank-1 recognition accuracy (\%) over years per data set.
In the following, we will discuss and highlight the best methodology and combination of training algorithm with loss function and optimizer to attain significant performance in PReID. 
\par
\textbf{Best Result per Dataset}
VIPeR is a small but challenging dataset. Therefore, mostly single models are utilized. The best result for VIPeR to date is shown by a single model i.e., 56.65. Whereas, the best result for VIPeR by Pairwise and Triplet model is 52.85 and 47.80. 
CUHK01 and CUHK03 are still challenging datasets. The best performance for these datasets is given by Pairwise models i.e., 93.10 and 94.90, respectively. Triplet models show second best result for these datasets, whereas Single models have reduced performance on these datasets. 
On i-LIDS, the single model shows the best result followed by Pairwise and then Triplet i.e., 88.0, 85.0 and 66.0, respectively. A single model showed the best result for PRID-2011 followed by Triplet model and then pairwise model i.e., 95.30, 81.0, and 70.0. 
CAVIAR is used only three times in which two times the model was Pairwise whereas, once it is evaluated over a single model. However, the best result is shown by Pairwise. 
On the other hand, MARS is also used three times, and each time it is evaluated with a single model. 
Market-1501 is the most evaluated dataset where the best result is shown by Single, Pairwise, and then Triplet model.     
\par
Finally, only~\cite{lin2017consistent} has evaluated their methodology on WARD dataset. The achieved Rank-1 rate for WARD dataset is $99.71\%$, which shows an ideal and almost optimal performance leaving little margin for future research. However, still in surveillance, one needs $100\%$ recognition rate to avoid the anomalies. 
\par

\textbf{Comparison} 
All solutions discussed here have trained their model with~\emph{SGD} with back-propagation algorithm. Majority of these works evaluated their models for PReID on CUHK03 (33), Market-1501 (29), VIPeR (27), and CUHK01 (26) datasets. Table. \ref{tab:cmc_firstRank} shows that VIPeR dataset is one of the most used in PReID problem since 2014, but it remains one of the most challenging datasets. One of the reason is its small size. However, future models need to be able to show good result either directly or through transfer learning.  
\par
Good performances have been shown in various large models; however, in real scenarios, the models need to be fast and effective. Almost in many video surveillance system, the concept of processing time is neglected for the sake to achieve higher accuracy. However, it should always be taken into consideration since it is very costly due to the requirement of powerful computers to run these deep models. Efforts have to be made in this regard to make methods more efficient and compatible for achieving high performance despite the smaller size of the network \cite{iodice2016strict}. The network can be reduced by either reducing the number of layers, number of parameters, or introducing a new scheme that has lower connectivity. In \cite{lavi_multiIETCV2018,lavi2016multi}, authors aimed at a trade-off between ranking accuracy and processing time by proposing a multi-stage ranking system and showed promising results.
\par
\textbf{Limitations and Future Directions:}
The task of PReID still suffers from the lack of training data samples. Although, this problem is addressed with the joint help of pairwise Siamese networks and data augmentation, and showed promising performances. However, utilizing such technique still has a major limitation to be applicable where it brings noises into the original data set that can effect the performance of the model in real-world scenarios.
Large scale datasets are needed to make models more reliable to tackle challenges such as pose and viewpoint variations in the images.  
In ML, a classification problem can be more often adopted to the problems with a limited number of classes in which a massive number of instances per class are highly demanded. To this end, the existing methods of ML, such as artificial neural networks allow solving classification problems with the limitations mentioned above. In PReID, some persons and the corresponding classes are increasing day by day. However, the number of instances acquired from camera networks is minimal. In this manner, PReID cannot be entirely taken in a position as a standard classification problem, particularly with DNNs. In contrary to a traditional classification problem, metric learning methods, as discussed in this paper, can help and overcome the limitation of deep models as an appropriate tool for solving PReID problem. 
\par
Many recent application areas such as autonomous-vehicles and aerial vehicles \cite{8099979}, \cite{Gaidon:Virtual:CVPR2016}, \cite{Smyth2018AVE} use synthetic data for training. Till date, no such dataset has been released for PReID problem. Using a game engine to generate and release a synthetic dataset for PreID can be a possible and viable solution. This could help PReID researchers in training the models and than using it in transfer learning over a smaller dataset. 
\par
The technique proposed for image-based PReID data set are not yet applied on video-based dataset in order to generate the sequence of target samples. This can be considered also as future direction on this research community.
\par
Cross-modality approach is also another hot research topic on PReID. This type of methods enables a PReID system to interact with other modalities to obtain alternative information about pedestrians. Those information can be further help the system to have a better analysis with hight performance accuracy under different scenarios. For instance, ~\cite{ye2018visible} proposes a cross-modality PReID for joint learning of thermal and visible domains, and addresses the issue of PReID in night-time. In this vein, domain knowledge transfer~\cite{narayan2019learning} is another interesting research line for PReID. Developing a system to learn specific knowledge (e.g., learning attributes) on a labeled dataset, and further evaluating it on an~\emph{unseen} data, which can make PReID system to be more deploy-able also in PReID open-set scenarios. 

\par
The discussed models in this paper mainly tried to address a short-term scenario by considering to camera views. Currently, to deploy a PReID system in a real-world over a long-term scenario is a daunting task. Also, a few works have considered to train the model in semi- and un-supervised manner. Such way of learning is more realistic to deploy a PReID system in a real-world scenario. However, the existing semi- and un-supervised methods have shown much weaker performance than supervised learning models which highlights that much work can be done in this area which can avoid the need of label data or can aid in generating label data. 
To this end, open-set PReID scenario~\cite{wang2016towards} addressed by a few works, particularly none deep learning-based methods. This is a challenging scenario which needs also to be considered more in order to accomplish the main goal of PReID. 
\par
Each model in Table~\ref{tab:cmc_firstRank} shows good performance for one or two benchmark datasets, but narrow to apply
to a realistic scenario of PReID due limitations as pointed above.
However, out of the 60 models, only one model (i.e., \cite{zhang2019scan}) shows optimal results for more than one database. This highlights the weakness in the current models and emphasizes researchers to come up with such models that can show good performance on at-least 50\% of the available datasets. On the other hand, a possible solution in the future research could be to propose specific rules/scenarios for combining all the datasets. Besides, rather individually releasing a new dataset, it will be good to add the new set of images to the old dataset and than evaluating their models. This will help the researchers to evaluate their model over a single dataset.
\par
An important factor to highlight is that while using DNN-based models, one has to take care of the size of the networks. DNNs have large number of parameters and the trained model require more disk space. Hence, when using pairwise or triplet models, these can result in a much heavier trained model. Eventually, it will be hard to store them on small embedded devices with limited memory. Therefore, models with fewer parameters and equal or better performance should be considered. 

\section{Final Remarks}
\label{sec:Conclusions}
Person re-identification is a challenging task for an intelligent video surveillance system with open application areas in numerous fields. Despite high importance it is still facing problems due to the poor performance of models in real-world scenarios. In this survey, we summarized recent advances with DNNs for the PReID task from 2014 to date. %
We have shown the type of models by taking into account of their implementation details for PReID. 
Besides, we highlighted all the available datasets in this domain. VIPeR dataset is the most challenging and widely used dataset available thus far. To handle the issue of lack of data, utilizing synthetic data is being proposed as a viable solution. 
%
Finally, it is essential to consider that besides enhancing the performance of the models, the size of the models (by reducing layers or number of parameters in the model) needs to be decreased without degrading the overall Rank-1 recognition rate. 

\section*{Acknowledgements}
This research was supported by S\~ao Paulo Research Foundation (FAPESP), under the thematic project "\emph{D\'ej\`aVu: Feature-Space-Time Coherence from Heterogeneous Data for Media Integrity Analytics and Interpretation of Events}" with grant number 18/05668-3.\newline 

\bibliographystyle{iet}
\bibliography{Parts/refs}

\end{document}